\documentclass[10pt,twocolumn,letterpaper]{article}

\usepackage[pagenumbers]{cvpr} 

\usepackage[dvipsnames]{xcolor}

\usepackage{algorithm}
\usepackage{algpseudocode}

\definecolor{cvprblue}{rgb}{0.21,0.49,0.74}
\usepackage[pagebackref,breaklinks,colorlinks,citecolor=cvprblue]{hyperref}

\def\paperID{7466} 
\def\confName{CVPR}
\def\confYear{2024}

\title{Context-PEFT: Efficient Multi-Modal, Multi-Task Fine-Tuning}

\author{Avelina Asada Hadji-Kyriacou\\
University of St Andrews\\
Department of Computer Science\\
{\tt\small lhk3@st-andrews.ac.uk}
\and
Ognjen Arandjelovic\\
University of St Andrews\\
Department of Computer Science\\
{\tt\small oa73@st-andrews.ac.uk}
}

\begin{document}
\maketitle
\begin{abstract}
This paper introduces a novel Parameter-Efficient Fine-Tuning (PEFT) framework for multi-modal, multi-task transfer learning with pre-trained language models. PEFT techniques such as LoRA, BitFit and IA\textsuperscript{3} have demonstrated comparable performance to full fine-tuning of pre-trained models for specific downstream tasks, all while demanding significantly fewer trainable parameters and reduced GPU memory consumption. However, in the context of multi-modal fine-tuning, the need for architectural modifications or full fine-tuning often becomes apparent. To address this we propose Context-PEFT, which learns different groups of adaptor parameters based on the token's domain. This approach enables LoRA-like weight injection without requiring additional architectural changes. Our method is evaluated on the COCO captioning task, where it outperforms full fine-tuning under similar data constraints while simultaneously offering a substantially more parameter-efficient and computationally economical solution.
\end{abstract}    
\section{Introduction}
\label{sec:intro}

Recent advancements in Parameter-Efficient Fine-Tuning (PEFT), exemplified by techniques such as LoRA, BitFit, and IA\textsuperscript{3} \cite{hu2021lora, zaken2022bitfit, liu2022fewshot}, have demonstrated their ability to deliver performance on par with full fine-tuning of pre-trained models for specific downstream tasks, all while significantly reducing the number of trainable parameters and GPU memory consumption. However, in the realm of multi-modal transfer-learning, there often arises a need for architectural adjustments -- such as Multi-Modal Causal Attention \cite{yao2023deepspeedvisualchat} -- or full fine-tuning -- as demonstrated by LLaVA \cite{liu2023visual}.

In response to these challenges we introduce a novel PEFT framework, Context-PEFT, which learns different groups of adaptor parameters based on the token's domain or purpose. This approach necessitates no further architectural modifications and is adaptable to various PEFT techniques. We conduct an extensive evaluation of our method on the COCO captioning tasks \cite{lin2015microsoft}, demonstrating its superior performance compared to full fine-tuning under similar data constraints, all while delivering a substantially more parameter-efficient and computationally economical solution.

We experiment with LoRA, BitFit and IA\textsuperscript{3} as adaptor methods for Context-PEFT and examine how they perform compared to full fine-tuning, no fine-tuning, as well as vanilla context-agnostic PEFT.

\section{Previous Work}
\subsection{Parameter-Efficient Fine-Tuning}
Large Language Models (LLMs) have been increasing in scale over recent years, with popular `baseline' models increasing in size from 350 million parameters in 2018 (BERT Large \cite{devlin2019bert}) to 7 billion parameters in 2023 (Llama 2 7B \cite{touvron2023llama}). With this increase in scale there has been an increase in popularity of PEFT techniques which have enabled the fine-tuning of large models using less compute and fewer resources. PEFT methods can be categorised by the ways they modify the underlying model architecture, which parameters are trainable and where new weights are introduced.

\paragraph{Adaptor Modules} MLP Adaptors \cite{rebuffi2017learning, houlsby2019parameterefficient} come in multiple variations but all share the same core idea of introducing a small residual fully-connected network after sub-layers in the transformer backbone. These adaptor layers are then trained while the original transformer layers remain frozen.

\paragraph{Soft Prompts} Soft prompt techniques -- such as Prefix-Tuning \cite{li2021prefixtuning, lester2021power} -- introduce learnable embeddings as a prefix to the input sequence. These embeddings can be trained as inputs to only the first layer and then propagated to all layers like normal tokens, or learned for each transformer sub-layer, replacing the results of embeddings of prior layers in the prefix region.

\paragraph{Selective Training} Selective training involves only training a subset of parameters. Such techniques include BitFit \cite{zaken2022bitfit} which trains only the bias terms of projection layers and LN-Tuning \cite{qi2022parameterefficient} which learn only the bias and scale parameters of the layer normalisation layers.

\paragraph{Additive Methods} Additive methods involve introducing learnable vectors that are applied point-wise to the activations in transformer sublayers. IA\textsuperscript{3} \cite{liu2022fewshot} applies an element-wise multiplication of learnable vectors to the queries and keys in attention layers as well as the intermediate activations of the feed-forward layers.

\paragraph{Reparameterization Methods} Another method to reduce the number of trainable parameters is by augmenting the original pre-trained weights using low-rank representations. One popular method is LoRA \cite{hu2021lora} which computes delta weights for the pre-trained projection matrices using low-rank matrix decompositions, $\delta W = A B$. Originally, LoRA was proposed for the key, query, value and output projections of attention layers, but the technique can also be applied to the FFN layers.

\subsection{Vision-Language Models}
Vision-Language models can often be classified under two categories: dual-encoder models which are more suited to classification and retrieval tasks \cite{zhai2022lit, radford2021learning, girdhar2023imagebind, jia2021scaling, singh2022flava}, and vision-encoder text-decoder models which are more suited to generative tasks such as captioning and VQA. Vision-encoder text-decoder models -- often refered to as Large Vision Language Models (LVLMs) -- can further be split into two categories based on their attention styles: causal-attention only models which incorporate image embeddings into the input context as projected tokens \cite{liu2023visual, yao2023deepspeedvisualchat, bai2023qwenvl}, and cross-attention augmented models which add additional attention layers to incorporate visual information \cite{alayrac2022flamingo}.

Among LVLMs, the causal-only style models have some advantages over cross-attention style models when augmenting pre-trained text-only models to accept other modalities. Cross-attention style LVLMs introduce a significant number of extra parameters which need to be trained, increasing both training compute requirement and inference latency. Care must also be taken when using cross-attention for multi-turn interleaved text and image input to ensure images are correctly localised to the corresponding text segments.

For causal LVLMs there are a multiple methods for adapting pre-trained text-only LLMs to accept visual inputs. DeepSpeed-VisualChat \cite{yao2023deepspeedvisualchat} opts to completely freeze the LLM and vision-encoder, training only the text embeddings and image projection layer. Additionally, the authors augment the attention mechanism such that two separate attention matrices are used for (1) text attending to text tokens and (2) text attending to visual tokens. They name this augmentation the `Multi-Modal Causal Attention Mechanism (MMCA),' with their intuition being that the attention weights of the two different modalities may interfere if jointly normalised. Alternatively, LLaVa \cite{liu2023visual} takes a two stage approach. Firstly the LLM and vision-encoder are frozen while the image projection weights are trained on a large image caption corpus. Next the LLM is unfrozen and is jointly fine-tuned with the image projections on a smaller language and vision multi-turn conversational dataset.

Both approaches have their own merits and their own drawbacks. The MMCA approach only trains a small subset of weights, but increases the compute requirements compared to vanilla attention using accelerated kernels such as FlashAttention \cite{dao2022flashattention}; as of writing, FlashAttention does not support explicit masks, which means the attention matrices for both modalities cannot be jointly computed using this acceleration structure and must instead fall back to either non-accelerated attention (which uses more GPU memory) or multiple rounds of FlashAttention (which increases training and inference latency). Since LLaVa uses vanilla causal attention it can take advantage of FlashAttention, but requires full fine-tuning of the LLM in the second stage of training.
\section{Our Approach}

We form a similar approach to that of LLaVa \cite{liu2023visual}, opting to fully freeze the vision-encoder and learn a linear projection to adapt image embeddings as inputs to the language model. Our method, however, differs in that we replace the full fine-tuning of the LLM with Context-PEFT and choose not to pre-train the linear projection on an auxiliary image-caption corpus as we want to train and evaluate our models on a standard dataset without any auxiliary training data.

\subsection{Model Selection}
The focus of this paper is efficiency, and with that in mind we have selected small and efficient models for this research.

For the LLM we wanted to use a Llama 2 style model with a similar size to GPT-2 small \cite{touvron2023llama, radford2019language}. However with Llama models starting at 7 billion parameters we opted to train our own model in house on 5 billion tokens from The Pile text corpus \cite{gao2020pile}. Details of the LLM are described in supplementary material Section~\ref{sec:llm}.

For the vision-encoder we opted to utilise the Swin Transformer V2 family of models due to their efficiency and size of produced embeddings \cite{liu2022swin}; for images of size $256 \times 256$ the Swin family produce 64 embeddings. Details of the chosen Swin models are detailed in supplementary material Section~\ref{sec:swin}.

\subsection{Image Inputs}
\begin{figure}
    \centering
    \includegraphics[width=\linewidth]{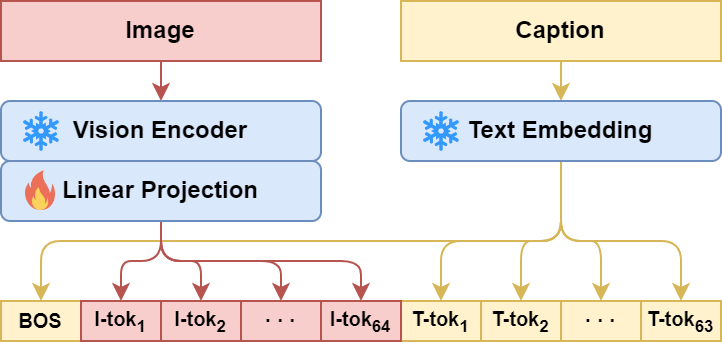}
    \caption{Image and text input structure. A pre-trained vision encoder produces image embeddings which are projected through a linear layer and concatenated with text embeddings of the corresponding caption.}
    \label{fig:input-seq}
\end{figure}

After passing the images through the vision-encoder the embeddings are projected to the LLM embedding dimension and added to the token sequence. The sequence is constructed such that it begins with a \texttt{[BOS]} token, followed by the 64 projected image embeddings, and up to 63 caption tokens which are post-padded by an \texttt{[EOS]} token and \texttt{[PAD]} tokens to achieve a fixed context length of 128. Almost all COCO captions fit into this 63 token budget with only a tiny faction of captions exceeding this length (and thus are truncated). This is illustrated in Figure~\ref{fig:input-seq}.

\subsection{Context-PEFT}
Our motivation behind Context-PEFT is inspired by Microsoft's introduction of MMCA; by decomposing the attention mechanism into two mode-specific sub mechanisms they change the way in which the text and image modalities interact and thus aggregate information. This is further exemplified by the fact that the authors found the addition of LoRA to have minimal effects on the results \cite{yao2023deepspeedvisualchat}, which further highlights the importance of controlling inter-modality interactions. Our approach is to instead directly modulate the weights applied to tokens of different modalities which we hypothesise would yield similar results to MMCA by modifying the embeddings passed to the attention mechanism rather than altering the mechanism itself. Our approach also has similarities to Mixture-of-Experts methods which have been shown to greatly improve performance of multi-domain tasks such as any-to-any machine translation \cite{shen2019mixture, gu2018universal}.

\begin{figure}
    \centering
    \includegraphics[width=\linewidth]{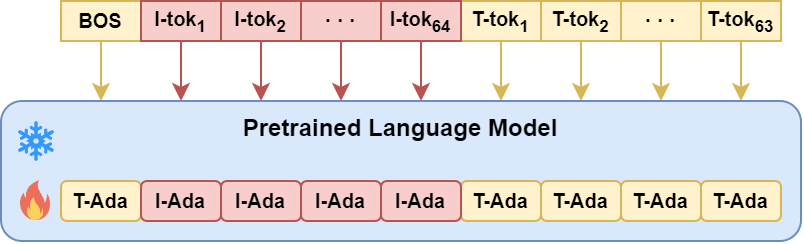}
    \caption{The overall model structure showing trainable adaptors used for different token types in the frozen language model.}
    \label{fig:model-structure}
\end{figure}

Each token in the sequence has a corresponding `context number' identifier which designates which adaptor weights should be used at each position of the sequence. When performing vanilla context-agnostic PEFT experiments we simply set all tokens to use the same context number. For all PEFT variants, context-specific and context-agnostic, we freeze all original model parameters and train only the injected adaptor parameters. The overall structure of the model is shown in Figure~\ref{fig:model-structure}.

Our choice of adaptor methods was motivated by their fine-tuning performance as reported in their original papers, ability to augment into context-specific forms, and ability to integrate into existing models with minimal modification. Although we opted to modify the code of our in-house model to facilitate easier debugging of adaptors during development, all adaptor methods can be incorporated into a PyTorch model using forward hooks without modifying any of the original code.

\paragraph{Context-LoRA}
We formulate a memory efficient way of computing context-specific LoRA such that the delta weight matrix is never fully materialised. To adapt any projection, $W$, for context LoRA with low rank matrices $A$ and $B$ the following algorithm is used:

\begin{algorithm}
    \caption{Memory efficient Context-LoRA algorithm}
    \begin{algorithmic}
        \State $x \gets states$
        \State $s \gets OneHot(contextNum)$
        \State $\delta h \gets Einsum( \text{`...ld,cdr,crD,...lc$\rightarrow$...lD'}, x, A, B, s )$
        \State $h \gets Wx + \delta h$
    \end{algorithmic}
\end{algorithm}

\begingroup
\setlength{\belowcaptionskip}{-24pt}
\setlength{\abovecaptionskip}{-6pt}
\begin{figure}[ht]
    \caption*{The letters in the einsum correspond to different input and output dimensions. $l$ corresponds to the sequence length dimension, $d$ and $D$ are the input and output dimensions of the projection, $r$ is the rank of the $A$ and $B$ decomposition matrices, and $c$ is the context selector dimension.}
\end{figure}
\endgroup

We apply context-LoRA to the query, key, value, and output projections in the attention layers as well as the up and down projections in the feed-forward layers.

\paragraph{Context-BitFit}
Traditionally, BitFit fine-tuning is performed by training only the bias terms; however this does not easily facilitate per-token bias terms. To combat this we reparameterize BitFit training by adding a secondary trainable bias term that depends on the token's context number. We apply BitFit to all projections in all layers of the transformer backbone.

\paragraph{Context-IA\textsuperscript{3}}
Our application of Context-IA\textsuperscript{3} is almost identical to Context-BitFit, with the main difference being element-wise multiplication of the projected states rather than addition. We apply IA\textsuperscript{3} to the key, value and intermediate FFN activations, as described by the original authors~\cite{liu2022fewshot}.
\section{Evaluation}
For most experiments we use caption perplexity (PPL) on the validation or test sets as a benchmark of model performance. When reporting results we the use weights from the epoch with best validation loss.

\subsection{Training}
Across all experiments we train using a batch size of 96 for eight epochs over the COCO dataset unless otherwise stated. We use the Adam optimizer with a fixed learning rate of $1e^{-4}$ and beta values of $(0.9, 0.95)$. We apply a dropout of 0.1 before the image projection layer, and a residual dropout after attention and feed-forward layers in the language model. We also apply flip, crop and colour augmentations to the images during training. For the loss function we use typical language modelling loss applied to the last 64 tokens of the sequence. Each experiment is conducted on a single RTX A4500 GPU and we make use of mixed precision training. We only use the training split of the COCO dataset to fine-tune our models, making no use of additional training data; this is to ensure repeatability and to explore the efficacy of context-PEFT in a data-constrained environment.

\begin{figure}[]
    \centering
    \includegraphics[width=\linewidth, trim={8mm 5mm 2mm 5mm}]{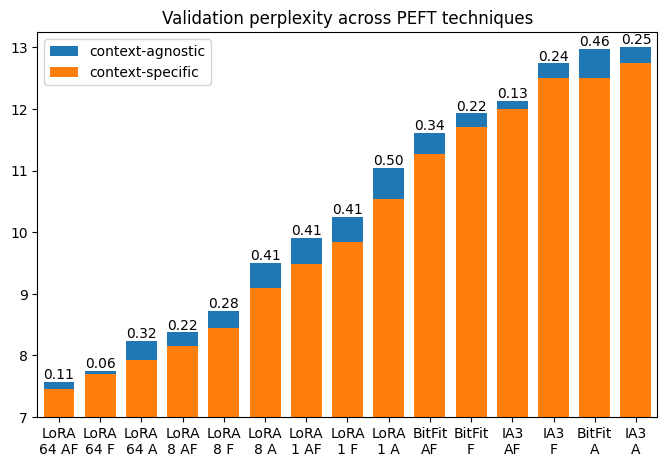}
    \caption{Validation perplexity after 8 epochs using the tiny Swin variant. `A' and `F' in the run name represent attention and feed-forward adaptation respectively, with the number representing the LoRA rank. The values above the bars represent the improvement in PPL for the context-specific variant.}
    \label{fig:fge-vppl}
\end{figure}

\subsection{Adapting Attention and Feed-Forward Layers}
\label{sec:fge-vppl}
To gain an understanding of how context-PEFT affects different parts of the transformer backbone we evaluate the performance of various PEFT methods when applying adaptation to both attention and FFN layers, and to either the attention or FFN layers. Our comparisons are conducted using the `tiny' version of the Swin transformer, where we analyze validation perplexity across both context-specific and context-agnostic versions of the three PEFT methods and while also varying the rank of the LoRA adaptors.

From the results summarised in Figure~\ref{fig:fge-vppl} we consistently find the context-specific adaptors outperforming their context-agnostic variant across all configurations, with the attention and feed-forward adaptor configurations performing best among each technique, and LoRA with a rank of 64 performing best overall.

When applying fine-tuning to a subset of layers (the `A' and `F' columns of Table~\ref{tab:fge-vppl}) we discover that adapting the feed-forward layers generally yields lower perplexity than adapting only the attention layers. However we observe the largest gains in perplexity for context-PEFT in the attention-only configurations compared to the context-agnostic variants which suggests that attention layers benefit the most from context-specific adaption, further validating our hypothesis of the importance of modality specific attention routing, but unlike DeepSpeed-VisualChat \cite{yao2023deepspeedvisualchat}, we are able to achieve this through context-PEFT alone. 

\begingroup
\setlength{\tabcolsep}{3pt}
\begin{table}[]
\resizebox{\linewidth}{!}{%
\begin{tabular}{l|ccc|ccc|ccc}
\hline
Adaptor & \multicolumn{3}{c|}{A} & \multicolumn{3}{c|}{F} & \multicolumn{3}{c}{AF} \\
\hline
IA3     & 13.01 & 12.75 & (-0.25)   & 12.74 & 12.51 & (-0.24)   & 12.13 & 12.0  & (-0.13) \\
BitFit  & 12.97 & 12.51 & (-0.46)   & 11.93 & 11.70 & (-0.22)   & 11.61 & 11.27 & (-0.34) \\
LoRA 1  & 11.04 & 10.54 & (-0.50)   & 10.25 & 9.83  & (-0.41)   & 9.91  & 9.49  & (-0.41) \\
LoRA 8  & 9.50  & 9.09  & (-0.41)   & 8.72  & 8.44  & (-0.28)   & 8.38  & 8.15  & (-0.22) \\
LoRA 64 & 8.24  & 7.92  & (-0.32)   & 7.75  & 7.69  & (-0.06)   & 7.57  & \textbf{7.46}  & (-0.11) \\
\hline    
\end{tabular}}
\caption{Validation perplexity after 8 epochs using the tiny Swin variant. `A' and `F' in the column name represent attention and feed-forward adaptation respectively. Context-agnostic PPL, context-specific PPL, and difference in PPL are represented by the left, center and right values in each column respectively.}
\label{tab:fge-vppl}
\end{table}
\endgroup

\begin{figure}[ht]
    \centering
    \includegraphics[width=\linewidth, trim={3mm 5mm 3mm 5mm}]{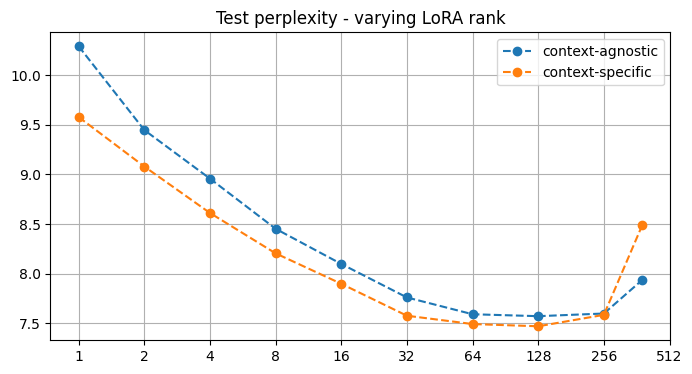}
    \caption{Test perplexity after 8 epochs using the tiny Swin variant, varying LoRA rank from 1 to 384 in the attention and feed-forward adaptor configuration.}
    \label{fig:lora-sweep}
\end{figure}

\subsection{Varying LoRA Rank}
\label{sec:lora-sweep}
Using the attention and feed-forward configuration with the `tiny' Swin model we sweep the LoRA rank from 1 to 384 (half of $d_{model}$) to evaluate how rank affects performance for both context-specific and context-agnostic variants.

From the results shown in Figure~\ref{fig:lora-sweep} it can be seen that test perplexity decreases as rank increases. However as the rank approaches 64 the perplexity plateaus and then sharply increases after a rank of 256, implying an optimal rank value between 64 and 128 for both context-specific and context-agnostic variants in our LLM. This suggests that the low-rank nature of LoRA has a beneficial regularising effect in data-constrained environments, such as our fine-tuning on the COCO dataset. This may also explain why full fine-tuning underperforms both in our experiments and some experiments carried out in the original LoRA paper \cite{hu2021lora}.

\begin{figure}[]
    \centering
    \includegraphics[width=\linewidth, trim={3mm 5mm 2mm 5mm}]{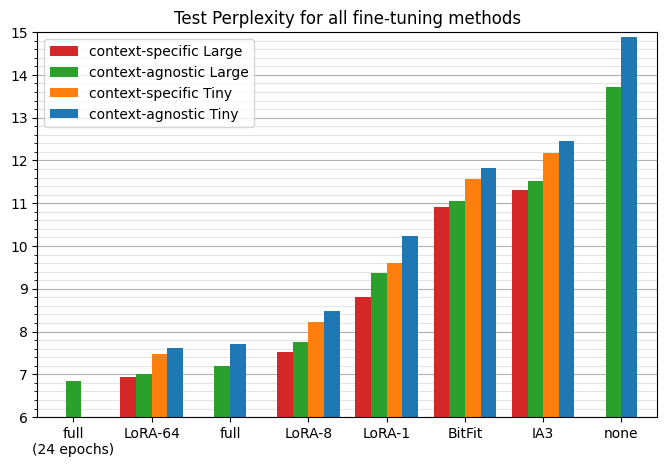}
    \caption{Test perplexity of different fine-tuning methods after 8
epochs for both vision encoder sizes.}
    \label{fig:vt-sweep}
\end{figure}

\begingroup
\setlength{\tabcolsep}{12pt}
\begin{table}[]
\resizebox{\linewidth}{!}{%
\begin{tabular}{l|cc|cc}
\hline
Adaptor             & \multicolumn{2}{c|}{Tiny} & \multicolumn{2}{c}{Large} \\
\hline
IA3                 & 12.46 & 12.18  & 11.52 & 11.32   \\
BitFit              & 11.83 & 11.56  & 11.04 & 10.90   \\
LoRA 1              & 10.23 & 9.61   & 9.38  & 8.82    \\
LoRA 8              & 8.48  & 8.22   & 7.76  & 7.53    \\
LoRA 64             & 7.61  & 7.48   & 7.00  & 6.93    \\
\hline
None                & 14.88 &        & 13.71 &         \\
Full FT             & 7.70  &        & 7.18  &         \\
Full FT (24 epochs) &       &        & 6.83  &         \\
\hline
\end{tabular}}
\caption{Test perplexity of different fine-tuning methods after 8 epochs for both vision encoder sizes. The left and right values correspond to context-agnostic and context-specific results respectively.}
\label{tab:vt-sweep}
\end{table}
\endgroup

\subsection{Effect of Vision Transformer Size}
\label{sec:vt-sweep}
We evaluate how vision encoder size affects test perplexity by comparing the `tiny' and `large' Swin models for a range of adaptors. We compare their performance for both the context-specific and context-agnostic adaptor variants in the `AF' configuration, with our results summarised in Figure~\ref{fig:vt-sweep} and Table~\ref{tab:vt-sweep}.

For each adaptor variant it can be seen that the large vision encoder with context-specific adaptation performs best, followed by the large vision encoder with context-agnostic adaption, both of which still perform better than the tiny vision encoder with context-specific adaptation. This is not surprising at all considering the large version of the Swin model has 7 times as more parameters ($195,202,932$) than the tiny version ($27,578,154$), which is a significantly larger increase in parameters than for context-specific vs context-agnostic adaption as shown in Table~\ref{tab:params}. This implies that higher quality image token embeddings are more important for image understanding in our models than the introduction of context-specific adaptation which only has twice as many trainable parameters than context-agnostic adaptation; despite this we can still see that context-specific adaptation can further lower perplexity which is especially evident in the lower rank LoRA adaptors.

Even though the Large swin transformer outperforms the Tiny swin transformer for each adaptor configuration in isolation, we can see that Context-LoRA-64 with Swin-Tiny outperforms all lower rank LoRA variants with Swin-Large; so even though Context-LoRA-64 has the most \textit{trainable} parameters, a full vision-text pipeline utilising Swin-Tiny + Context-LoRA-64 has fewer \textit{total} parameters and lower perplexity than Context-LoRA-8 + Swin-Large. This has important implications for scenarios where model size may be more important than model performance, for example when performing inference on embedded systems with resource limitations preventing the use of a larger vision encoder.

\begingroup
\setlength{\tabcolsep}{18pt}
\begin{table}[]
\resizebox{\linewidth}{!}{%
\begin{tabular}{l|c|c}
\hline
Adaptor & \multicolumn{1}{c}{Context-Agnostic} & \multicolumn{1}{|c}{Context-Specific} \\
\hline
IA3     & 55,296                               & 110,592                              \\
BitFit  & 119,808                              & 239,616                              \\
LoRA-1  & 202,752                              & 405,504                              \\
LoRA-8  & 1,622,016                            & 3,244,032                            \\
LoRA-64 & 12,976,128                           & 25,952,256                           \\
\hline
Full FT & \multicolumn{2}{c}{153,196,032}                                             \\
\hline
\end{tabular}}
\caption{The number of trainable parameters in the LLM backbone for different adaptor variants in the attention and feed-forward configuration.}
\label{tab:params}
\end{table}
\endgroup

We can also see that LoRA with a rank of 64 outperforms full fine-tuning for the respective vision transformer sizes when given the same data budget of 8 epochs, both with and without context-specific adaptation. We do an additional run and find it takes 24 epochs of full fine-tuning to outperform context-LoRA with the large Swin model as the vision encoder.

\subsection{Attention Map Observations}
\label{sec:att-map}

The vision transformers we used were pre-trained for image classification on ImageNet-1k (with the large version trained on ImageNet-22k before finetuned for ImageNet-1k), meaning it may be reasonable to assume the captioning abilities of our models come purely from the image classification abilities of the vision-encoder, rather than the language model's ability to extract and reason with the information in the image tokens themselves. However, we show this is not the case by analysing the attention weights of our trained models.

\begin{figure}[b]
    \centering

    \begin{subfigure}{0.495\linewidth}
        \includegraphics[width=\linewidth]{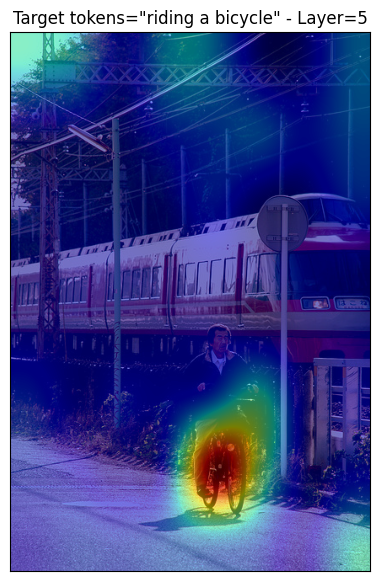}
        \caption{}
        \label{fig:attention-heatmaps-a}
    \end{subfigure}
    \begin{subfigure}{0.495\linewidth}
        \includegraphics[width=\linewidth]{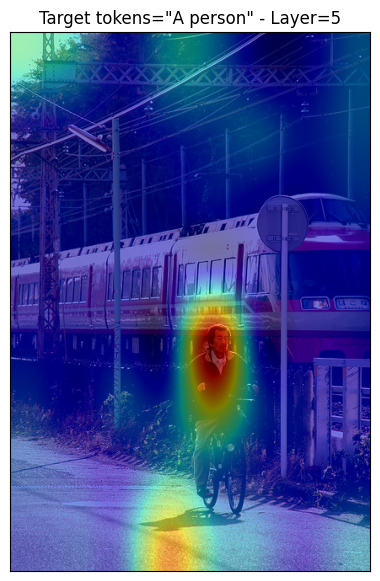}
        \caption{}
        \label{fig:attention-heatmaps-b}
    \end{subfigure}

    \begin{subfigure}{0.495\linewidth}
        \includegraphics[width=\linewidth]{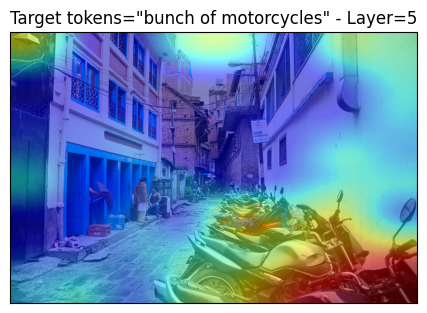}
        \caption{}
    \end{subfigure}
    \begin{subfigure}{0.495\linewidth}
        \includegraphics[width=\linewidth]{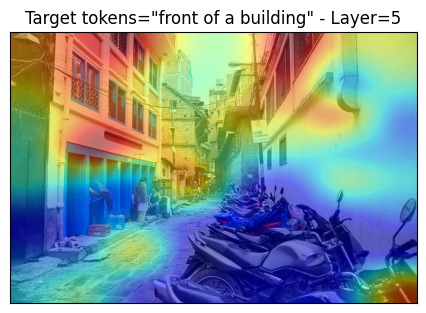}
        \caption{}
    \end{subfigure}

    \begin{subfigure}{0.495\linewidth}
        \includegraphics[width=\linewidth]{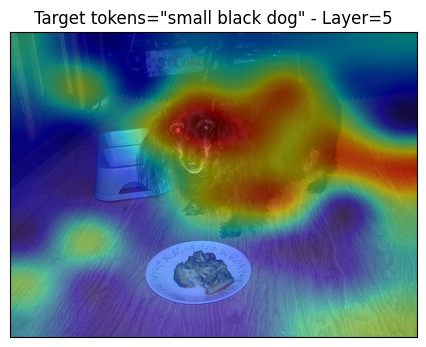}
        \caption{}
    \end{subfigure}
    \begin{subfigure}{0.495\linewidth}
        \includegraphics[width=\linewidth]{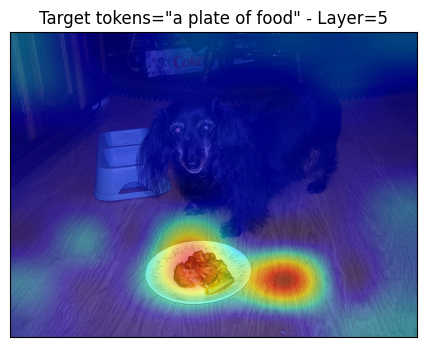}
        \caption{}
    \end{subfigure}
    
    \caption{Heatmap visualisations depicting image token attention weights for spans of caption tokens.}
    \label{fig:attention-heatmaps}
\end{figure}

To collect these attention heatmaps we take image-caption pairs from the the COCO test set, forward pass them through our trained Context-LoRA-64-AF model and observe the attention weights for spans of caption tokens to gain an understanding of the importance of different parts of the image for different parts of the caption. More specifically, we take the attention weight matrix for layer $n$ and slice the resulting matrix such that we only keep columns corresponding to the image tokens and rows corresponding to the desired span of caption tokens. Finally we sum across the query and head dimension and reshape the resulting vector into an $8 \times 8$ matrix which we upscale and overlay on the original image using a colour gradient.

From examples given in Figure~\ref{fig:attention-heatmaps} and supplementary Section~\ref{sec:heatmaps} we observe that the text tokens place attention on specific regions of the images which have clear semantic relevance with surprisingly fine granularity; for example, we observe that the model is able to tell the difference between a bicycle and the person riding on it, as shown in Figures~\ref{fig:attention-heatmaps-a} and \ref{fig:attention-heatmaps-b}. This shows that -- even though we train on image segmentation data -- the model could generalise to perform panoptic segmentation by learning linear classifiers on the attention maps.

We also notice that the first and last image tokens (top-left and bottom-right image patches) occasionally have unusually high attention weights. The large top-left weighting may be explained through the attention sinking phenomenon~\cite{xiao2023efficient}, while the bottom-right weighting can be explained by the fact that the last image token is also the location that produces the first text-token in auto-regressive text generation. We hypothesise that both phenomena would disappear if we wrapped the image embeddings between additional `beginning' and `end' tokens which may further lead to performance improvements through more salient attention behavior.

\section{Discussion}

\subsection{Applications and Limitations}
The primary application for Context-PEFT is in compute and data-constrained environments; for example when there is a low volume of training data, when there are GPU memory restrictions necessitating the use of smaller models and/or fewer trainable parameters, when there is limited training time which may prevent full fine-tuning from converging to a good minima, or when models must be deployed for inference on low-resource hardware such as consumer grade workstations, mobile devices and embedded systems.

However, if the aforementioned constraints are not a concern, it is unlikely Context-PEFT will outperform full fine-tuning. This is likely due to the fact that the embedding spaces of SOTA language models are often wide enough to represent multiple modalities without the need for context-specific weights when trained to convergence on a large volume of training data, as shown by Qwen-VL and LLaVa \cite{bai2023qwenvl, liu2023visual}.

\subsection{Future Work}
There are numerous improvements that can be integrated from concurrent and previous works to further push the performance of Context-PEFT as a fine-tuning method. These methods and some novel suggestions for future work are summarized below:

\paragraph{Exploring More PEFT Techniques} Further exploration of PEFT techniques may also yield greater fine-tuning performance at potentially lower budgets. One such example would be investigating how context-specific scaling and offset parameters affect the performance of LN-tuning. Another avenue worth exploring is combining PEFT techniques, such as combining LoRA and BitFit to not only perform low rank updates of linear layers but also update their corresponding bias terms.

\paragraph{Improved Vision Encoder} From our evaluation in Section~\ref{sec:vt-sweep} we have shown that the quality of input embeddings for other modalities can greatly affect the performance of the fine-tuned language model. There are many vision encoder alternatives:
\begin{itemize} 
    \item \textbf{CLIP-ViT}~\cite{radford2021learning}, which was trained specifically to connect vision and textual content in addition to producing a larger number of image embeddings than the Swin Transformer family.
    
    \item \textbf{VQ-VAE}~\cite{oord2018neural}, which produces a greater number of image embeddings than ViT. Additionally, instead of using an image projector to align the latent space for the language model, we can instead take advantage of the discrete nature of the latent space by learning a set of embeddings that map from the VAE `vocabulary' directly into the embedding space of the language model.
    
    \item \textbf{VQ-GAN}~\cite{esser2021taming}, which has the same benefits as VQ-VAE with the addition of producing latents which were also trained with conditioning on textual features, especially if the VQ-GAN/VQ-VAE was fine-tuned for use with latent diffusion model~\cite{rombach2022highresolution}.
    
    \item \textbf{DETR}~\cite{carion2020endtoend} object detection encoder, which may allow the language model to learn better semantic connections between text tokens and image tokens.
    
    \item \textbf{Encoder Fine-Tuning} as in Qwen-VL~\cite{bai2023qwenvl}, where a pre-trained Vision Transformer is used a starting point and then further fine-tuned with a Visual Question Answering (VQA) objective.

    \item \textbf{Encoder PEFT}, similar to Qwen-VL, it is worth exploring the use of PEFT techniques like LoRA to fine-tune the vision encoder without performing full parameter updates.
\end{itemize}

\paragraph{Improved Pre-Training} As shown in the LLaVa paper~\cite{liu2023visual}, pre-training on a mass image-caption corpus improves the alignment of the image projector which may have similar effect to using an improved vision encoder. CC3M and LAION-COCO are two such corpora which are well suited to pre-training the image-projector.

\paragraph{Improved Image Projection} Replacing the linear projector with an MLP has also been shown to improve alignment quality, as explored in LLaVa-1.5~\cite{liu2023improved}. It is also worth investigating the addition of learnt absolute position embeddings after the image projection as the some vision encoder architectures, such as CNNs, do not produce embeddings which contain positional information.

\paragraph{Improved Base Language Model} It is well known that larger language models have improved zero-shot generative capabilities over smaller language models \cite{brown2020language}, so an obvious choice for future work is to apply our framework to a larger base language model as this should also lead to higher quality caption generation.

\paragraph{Exploring Hyperparameters} Optimal hyperparameter choice can often improve the quality of a model without changing the model architecture or training objective itself. It is therefor worth exploring the impact of optimizer choice (exploring alternative such as AdamW, Sophia, Adabelief, Lookahead, Sharpness-Aware-Minimization), optimizer hyperparemeters, learning rate schedules, and the impact of dropout when applied to different subsets of layers and with different drop-rates.

\paragraph{Pruning and Quantization} Context-PEFT (and PEFT techniques in general) are often lightweight for both training and inference, but these techniques can further be combined with weight quantization and pruning to further reduce model footprint with minimal effect on performance~\cite{chen2020lottery, dettmers2023qlora}.

\paragraph{Training Objectives} In our work we made use of only the causal language modelling loss as our training objective, but there are numerous other choices which may lead to a higher quality model. One example of a training objective that may improve model quality is causal masked language modelling, where a span of tokens in the caption are replaced with a single \texttt{[mask]} token and the model is tasked with predicting the string of masked tokens; this objective may also condition the model for down-stream tasks such as visual question answering due to its similarity with the training objective. Another training objective worth exploring would be online reinforcement learning where the reward signal for generated sequences is given by CLIP alignment score~\cite{radford2021learning}.

\paragraph{Other Modalities} We believe that Context-PEFT can be extended to many other modalities and potentially allow the fusion of several modalities into a unified model. Examples of a multi-modal tasks suitable for Context-PEFT include processing audio for speech recognition, and processing video for Video Question Answering.

\paragraph{Boundary Tokens} As shown in Section~\ref{sec:att-map} there are some unusual attention behaviours at the boundaries between different modalities. We hypothesize that it may be possible to alleviate these phenomena by wrapping each modality segment with special tokens, \texttt{[section]} and \texttt{[/section]}. The introduction of these section boundary tokens also opens up opportunities for auxiliary training losses by treating the end of section token as a \texttt{[CLS]} token~\cite{devlin2019bert}, which would facilitate the learning of linear classification heads for a wider variety of downstream tasks.

\paragraph{Prompt Injection Mitigation} Prompt injection attacks are typically mitigated through cleansing incoming messages and using special formatting to prevent exploits such as the model interpreting user inputs as system messages. However, bad actors often still find ways to overcome these protections and bypass the safety guardrails of publicly available LLMs-as-a-service. We hypothesize that Context-PEFT could potentially be used to mitigate prompt injection by using different adaptors for system messages, user inputs and generated responses, which -- when combined with reinforcement learning as is typically used for `chat' models -- may condition the model to completely ignore any attempts to inject harmful prompts within user input segments. 

\begin{figure}[ht]
    \centering
    \includegraphics[width=\linewidth]{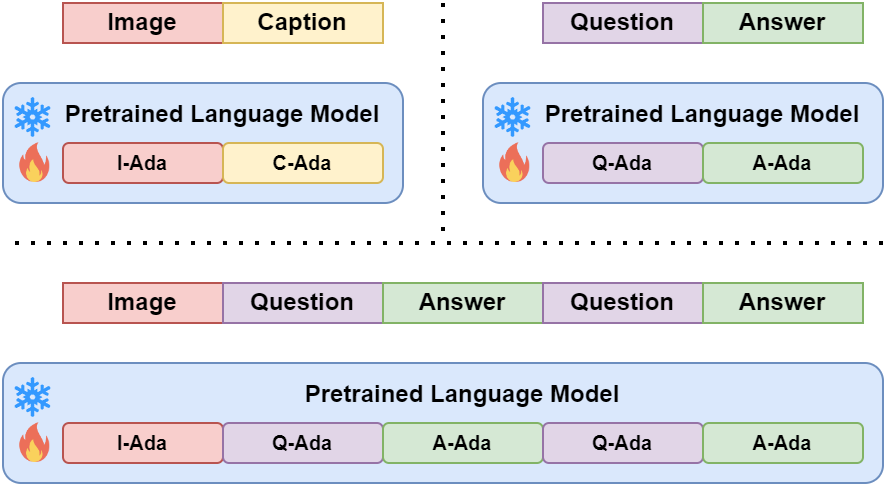}
    \caption{Parallel training for VQA with Context-PEFT.}
    \label{fig:parallel-training}
\end{figure}

\paragraph{Parallel Training} We also believe that Context-PEFT has untapped potential for parallel training on different tasks to then assemble the learnt adaptors into a unified model for further fine-tuning on a combined task. One example use case -- illustrated in Figure~\ref{fig:parallel-training} -- is training one set of context-PEFT adaptors for image captioning, training another set of context-PEFT adaptors for textual question answering (where questions and answers represent different contexts), and then fine-tuning the unified ensemble of adaptors for a downstream Visual Question Answering task. Taking this to the extreme, we could train context-PEFT on dozens of tasks and modalities in parallel, effectively treating each set of adaptors as `plugins' which can be integrated into a single unified model.

\subsection{Conclusion}
In this paper, we have introduced Context-PEFT as a multi-modal, multi-task fine-tuning framework to efficiently adapt pre-trained language models to other modalities in a data- and parameter-efficient manor by learning multiple groups of adaptor parameters which are specific to each token's domain.

Through extensive experimentation with the COCO dataset and captioning task we validate the potential of Context-PEFT, evaluating our framework for a variety of adaptor methods and fine-tuning configurations, and we show our method is very competitive with full fine-tuning for data-constrained and compute-limited environments.

Finally, we suggest a range of possible applications for context-PEFT and provide a variety of future research directions and potential improvements to further explore its performance and capabilities beyond our use-case of image captioning.
{
    \small
    \bibliographystyle{ieeenat_fullname}
    \bibliography{main}
}

\clearpage
\setcounter{page}{1}
\maketitlesupplementary

\section{Vision Encoder Details}
\label{sec:swin}

We experiment with two versions of the Swin Transformer V2 \cite{liu2022swin}, `Tiny' and `Large'.

The tiny variation is used for initial experimentation and fine grained comparisons of different adaptors. This variation was pre-trained for image classification on ImageNet-1k \cite{russakovsky2015imagenet} and has a final embedding dimension of 768. The weights and model were provided by HuggingFace\footnote{\href{https://huggingface.co/microsoft/swinv2-tiny-patch4-window8-256}{SwinT V2 Tiny on Hugging Face}}.

The large variation is used for final evaluation and coarse grained comparisons with the best adaptor variants. This variation was pre-trained for image classification on ImageNet-21k \cite{russakovsky2015imagenet}, further fine-tuned for ImageNet-1k, and has a final embedding dimension of 1536. The weights and model were provided by HuggingFace\footnote{\href{https://huggingface.co/microsoft/swinv2-large-patch4-window12to16-192to256-22kto1k-ft}{SwinT V2 Large on Hugging Face}}.

\section{LLM Pre-Training Details}
\label{sec:llm}

The LLM has 150 million parameters with an embedding dimension of 768, 12 layers of transformer blocks, 12 attention heads per attention layer utilising Rotary Position Embeddings \cite{su2022roformer} with an adjusted base frequency (RoPE ABF \cite{xiong2023effective}), SwiGLU activation \cite{chowdhery2022palm} in the feed forward layers with an intermediate size of 6144/3072, and using a frozen embedding layer taken from OPT-125m\footnote{\href{https://huggingface.co/facebook/opt-125m}{OPT-125m on Hugging Face}} to speed up convergence. We make use of FlashAttention 2 \cite{dao2022flashattention} to attain better performance and memory utilisation than vanilla dot-product attention.

We also train the model with a Transformer-XL style cache and shifting window \cite{dai2019transformerxl}, although instead of saving keys and values we cache the hidden states pre-projection and recompute them for each segment. This introduces some additional training and inference latency when the cache is in use, but comes with the advantage of generating partial gradients for the XL-cache and halves the persistent memory cost for the cache when accumulating gradients over several sub-batches.

The Sophia optimiser was used \cite{liu2023sophia} for pre-training, with a warm-up of approximately 2000 batches to a learning rate of 6e-4 and cosine annealed to a learning rate of 6e-5, using a batch size of 480, sequence length of 1024 tokens (with additional XL memory of 1024 states for total context size of 2048) and optimiser hyper-parameters suggested by the original Sophia authors.

We train for 5 billion tokens on the now defunct Pile dataset. The Pile was removed from its hosting platform due to containing some copyrighted materials, but this removal occurred after the start of this work. Alternatives such as the Pile-Uncopyrighted\footnote{\href{https://huggingface.co/datasets/monology/pile-uncopyrighted}{Pile-Uncopyrighted on Hugging Face}} exist and should provide comparable performance in terms of pre-training.

\section{Heatmaps}
\label{sec:heatmaps}

We provide auxiliary heatmap visualisations to further illustrate the image-text semantic understanding capabilities of our model.

\begin{figure}[ht]
    \centering
        \includegraphics[width=\linewidth]{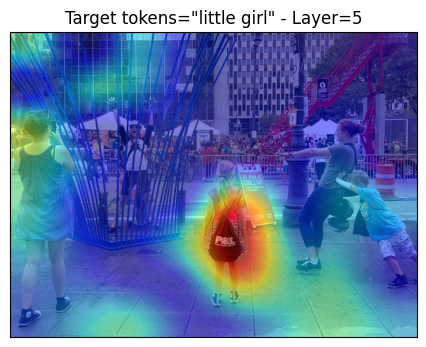}
\end{figure}

\begin{figure}[ht]
    \centering
        \includegraphics[width=\linewidth]{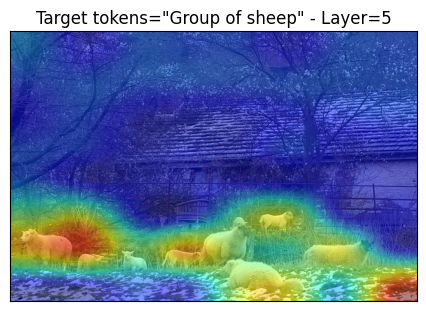}
\end{figure}

\begin{figure}[ht]
    \centering
        \includegraphics[width=\linewidth]{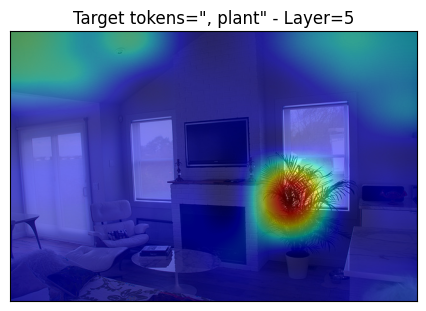}
\end{figure}

\begin{figure}[ht]
    \centering
        \includegraphics[width=\linewidth]{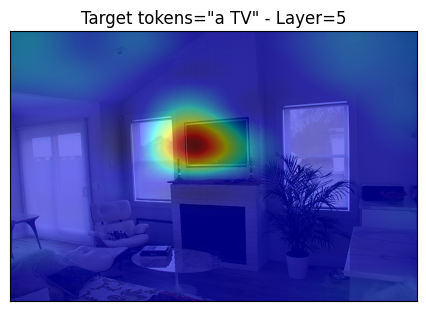}
\end{figure}

\begin{figure}[ht]
    \centering
        \includegraphics[width=\linewidth]{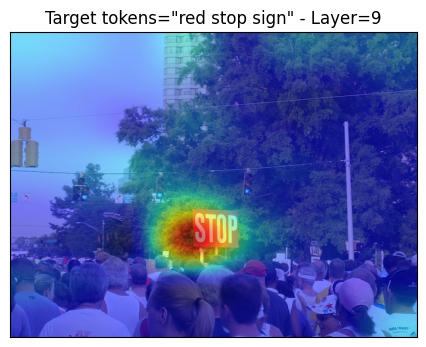}
\end{figure}

\begin{figure}[ht]
    \centering
        \includegraphics[width=\linewidth]{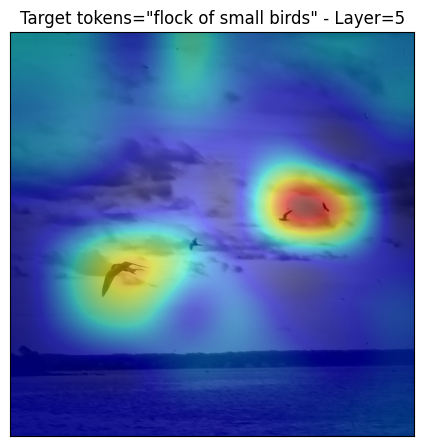}
\end{figure}

\begin{figure}[ht]
    \centering
        \includegraphics[width=\linewidth]{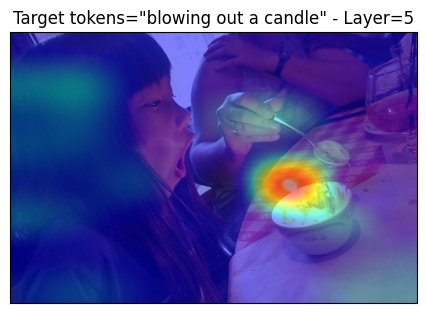}
\end{figure}

\begin{figure}[ht]
    \centering
        \includegraphics[width=\linewidth]{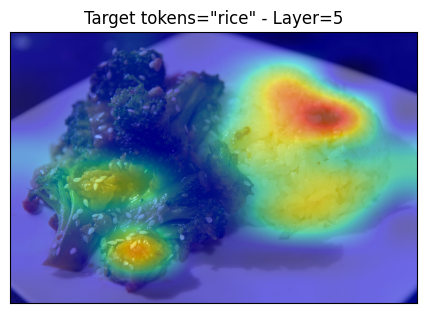}
\end{figure}

\begin{figure}[ht]
    \centering
        \includegraphics[width=\linewidth]{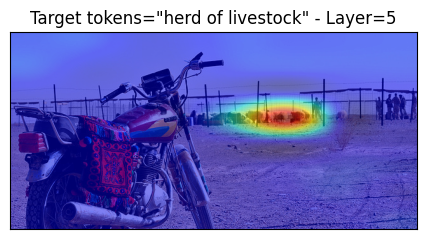}
\end{figure}

\begin{figure}[ht]
    \centering
        \includegraphics[width=\linewidth]{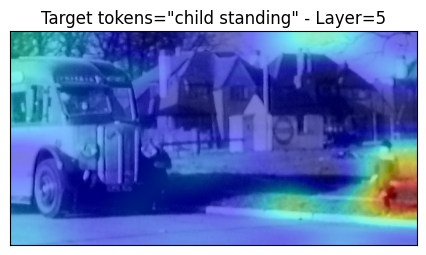}
\end{figure}

\begin{figure}[ht]
    \centering
        \includegraphics[width=\linewidth]{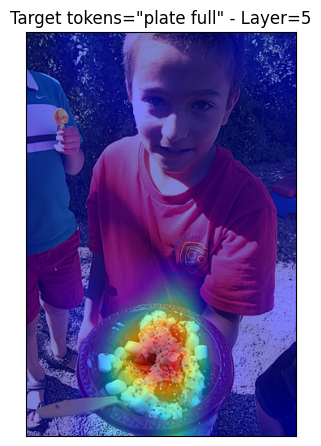}
\end{figure}

\begin{figure}[ht]
    \centering
        \includegraphics[width=\linewidth]{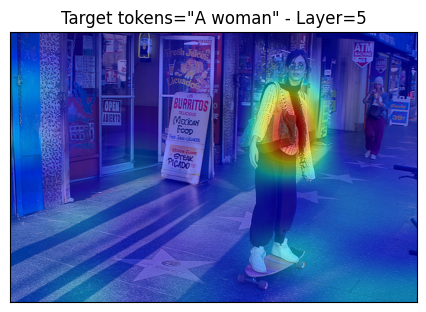}
\end{figure}

\begin{figure}[ht]
    \centering
        \includegraphics[width=0.666\linewidth]{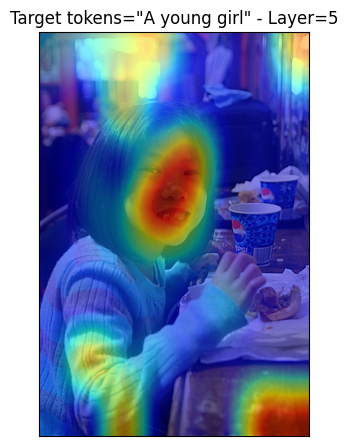}
\end{figure}

\begin{figure}[ht]
    \centering
        \includegraphics[width=0.875\linewidth]{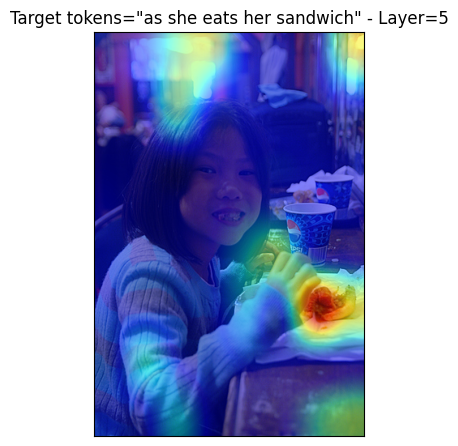}
\end{figure}

\end{document}